%% file: root.tex
\documentclass[letterpaper, 10 pt, conference]{ieeeconf}  

\IEEEoverridecommandlockouts                              

\input{custom}

\title{\LARGE \bf
A Preview of Open-Loop and Feedback Nash Trajectories \\in Racing Scenarios*
}

\author{Matthias Rowold$^{1}$
\thanks{*This work was not supported by any organization}
\thanks{$^{1}$Matthias Rowold is with the Chair of Automatic Control, Department of Mechanical Engineering, TUM School of Engineering and Design, Technical University of Munich, 85748 Garching
        {\tt\small matthias.rowold@tum.de}}%
}

\begin{document}

\maketitle
\thispagestyle{empty}
\pagestyle{empty}

\begin{abstract}
Trajectory planning for autonomous race cars poses special challenges due to the highly interactive and competitive environment. Prior work has applied game theory as it provides equilibria for such non-cooperative dynamic problems. This contribution introduces a framework to assess the suitability of the Nash equilibrium for racing scenarios. To achieve this, we employ a variant of \acs{iLQR}, called \acs{iLQGame}, to find trajectories that satisfy the equilibrium conditions for a \acl{LQ} approximation of the original game. In particular, we are interested in the difference between the behavioral outcomes of the open-loop and the feedback Nash equilibria and show how \acs{iLQGame} can generate both types of equilibria. We provide an overview of open problems and upcoming research, including convergence properties of \acs{iLQGame} in racing games, cost function parameterization, and moving horizon implementations.
\end{abstract}

\section{Introduction}
The complexity of considering interactions between autonomous vehicles and their interactions with human agents presents a significant challenge in trajectory planning. In established sequential methods, the autonomous vehicle of interest -- hereafter referred to as the ego vehicle -- is concerned with predicting the most likely trajectories of all relevant agents to react with a collision-free trajectory. This is often sufficient, as applications have shown. However, sequential approaches neglect the reciprocal nature of scenarios, meaning the other agents respond to the executed motion of the ego vehicle, creating a bidirectional interdependence. Planning approaches that consider and incorporate this mutual dependency are categorized as interaction-aware.

Interaction-aware approaches promise to generate trajectories with a lesser degree of conservatism compared to sequential approaches. This means they are performant and human-like even in environments with rapidly increasing prediction uncertainties that, under a sequential approach, would lead to overly cautious trajectories. 
By leveraging the knowledge that other agents react to the ego vehicle, including collision avoidance, interaction-aware approaches can influence the other agents' behaviors to a certain extent to achieve more progressive and less risk-averse behaviors. An application-oriented goal of interaction-aware planning is to generate behaviors that are seamlessly integrateable into traffic scenarios like lane changes \cite{Schmidt2019}, ramp merges \cite{LeCleach2022}, or crosswalks \cite{FridovichKeil2020, Crosato2023}. In addition to traffic, autonomous racing is another domain that heavily relies on interactions. In racing, strategies like overtaking, blocking, and faking are common, requiring anticipating the opponent's reaction to the ego-trajectory to be successful and safe. A major distinction to traffic scenarios is that the desired behavior in racing is usually competitive, i.e., non-cooperative.

Interaction-aware planning approaches employ multi-agent planning with a joint cost function, \aclp{POMDP}, reinforcement learning, and game-theoretical concepts. The latter seems especially fitting for autonomous racing since game theory provides concepts for non-cooperative behaviors in environments where the agents cannot communicate. Furthermore, the objective of each agent is similar and known. The goal is to maximize speed and be ahead of the opponents, in contrast to traffic scenarios with a wide range of objectives. Given that game-theoretic concepts require assumptions about the cost functions that govern the agent's decisions, their use for racing seems appropriate.

In this and the following work, we will analyze the suitability of a game-theoretic concept, the Nash equilibrium, for trajectory planning in autonomous racing. Our focus lies on two types of Nash equilibria: the open-loop and the feedback equilibria. Both types have been considered in previous work, but they have not been compared regarding their behavioral outcomes.

\subsection{Related Work}
\label{sec:related_work}
Most game-theoretic planning approaches in traffic and racing scenarios are concerned with finding trajectories that fulfill the requirements of a Nash equilibrium. At a Nash equilibrium, no agent, in the following called player, has an incentive to alter its strategy unilaterally. Depending on the information structure of the formulated game, one obtains either the open-loop or the feedback solution. Each player has to commit to a sequence of control inputs at the beginning of the game for an open-loop solution. In contrast, for a feedback solution, the players look for strategies that allow them to react to the current state in each stage of the game. A more detailed introduction to these concepts will follow in Section \ref{sec:problem_formulation}.

We categorize the existing approaches for game-theoretic trajectory planning into the following groups:
\subsubsection{Offline policy generation}
Fisac et al. \cite{Fisac2019} discretize the state space and determine the optimal policy for a feedback Stackelberg equilibrium offline via dynamic programming. This policy can be applied efficiently only, but the offline calculations suffer from the curse of dimensionality, so only a few players and coarse discretizations are possible.
Bhargav et al. \cite{Bhargav2021} perform extensive offline computations as well. However, they do not solve for equilibria, but policies with a high probability of successful overtaking for different race track positions.

Zheng et al. \cite{Zheng2022} formulate racing as a two-player zero-sum game in extensive form and determine the optimal strategy via \ac{CFR} minimization.  

\subsubsection{Sampling-based}
Liniger and Lygeros \cite{Liniger2020} formulate bi-matrix games by sampling trajectory candidates for two players. Solving these games for a Nash equilibrium results in open-loop trajectories. Feedback is introduced when the planning is performed with a receding horizon.

\subsubsection{\Ac{IBR}}
In \ac{IBR} approaches, the players optimize their trajectories alternately while keeping all other players' trajectories fixed. If this algorithm converges, no player is incentivized to alter its decision, making it a Nash equilibrium. Sensitivity-enhanced algorithms have been proposed in \cite{Wang2021, Wang.2019, Spica.2020} for drone and vehicle racing. Since the trajectories are optimized as a whole, the result is an open-loop equilibrium.

\subsubsection{\Ac{DDP}}
\ac{DDP} \cite{MAYNE1966} is a trajectory optimization method that iteratively performs backward- and forward passes to refine the trajectory. During the backward pass, an incremental control law is generated based on second-order approximations of the cost and dynamics along a nominal trajectory. The forward pass updates the nominal trajectory based on the incremental control law. Using only a first-order approximation of the dynamics results in \ac{iLQR} \cite{Li2004}.

Fridovich-Keil et al. \cite{FridovichKeil2020} transfer this iterative procedure to dynamic games. They approximate each player's cost function with a second-order tailor expansion and linearize the dynamics. The result is a linear quadratic game for which -- like for time-discrete \acp{LQR} -- analytic solutions exist \cite{Basar1999}. If this algorithm, called \ac{iLQGame}, converges, a Nash equilibrium to a local approximation of the game is found. Since \ac{iLQGame} provides feedback strategies for each player, the solution constitutes a feedback Nash equilibrium.
Similarly, Schwarting et al. \cite{Schwarting2021} solve a quadratic game in the backward pass to compute incremental feedback laws for the players. However, they plan in belief space, making it a multi-player variant of \ac{iLQG} control. 

Kavuncu et al. \cite{Kavuncu.2021} show that their used cost function constitutes a potential game so that the problem can be reformulated as a conventional \ac{OCP}. Using \ac{iLQR} to solve the \ac{OCP}, they find an open-loop Nash equilibrium.

\subsubsection{First-order optimality condition}
ALGames by Le Cleac'h et al. \cite{LeCleach2022} solve a root-finding problem to fulfill the first-order optimality condition of a Nash equilibrium. They enforce constraints with an augmented Lagrangian method and obtain a local open-loop Nash equilibrium with reported superior computation times compared to \ac{iLQGame}.
Zhu and Borrelli \cite{Zhu.2023} develop an \ac{SQP} variant to find a Nash equilibrium as a solution to the \ac{KKT} conditions. As in \cite{LeCleach2022}, the algorithm, if it converges, finds an open-loop equilibrium.

\section{Scope}
Some of the above approaches are compared regarding their calculation times \cite{LeCleach2022} and convergence success rates \cite{Zhu.2023}. We, however, are interested in their behavioral outcome and performance in racing scenarios. We focus on the comparison of open-loop and feedback solutions since the two types of equilibria can lead to entirely different solutions, as shown in Starr and Ho \cite{Starr1969b}.

With this contribution, we propose a framework to assess both concepts in racing scenarios. We identify the \ac{iLQGame} approach as a suitable method for finding open-loop and feedback Nash equilibria. Although only the feedback case is analyzed in \cite{FridovichKeil2020}, an adaption allows the approach to find open-loop equilibria. The adaption does not require altering the cost function or changing the fundamental working of the algorithm so that differences in the solutions due to different cost functions can be ruled out. This ensures the comparability of different solutions caused by the type of equilibrium.

In the following, we will first provide game-theoretic preliminaries and introduce the two types of equilibria. Section \ref{sec:solving_game} explains the \ac{iLQGame} algorithm, and Section \ref{sec:dynamics_cost} formulates our racing game with its dynamics and the players' cost functions. In Section \ref{sec:results}, we will show exemplary results of open-loop and feedback trajectories to illustrate the necessity of a more detailed examination.

\section{Game-Theoretic Preliminaries}
\label{sec:problem_formulation}
The dynamics describing the propagation of the joint state $\bm{x}_k$ for a dynamic game with $N$ players is given by:
\begin{equation}
	\label{eq:game_nonlin_dynamics}
	\bm{x}_{k+1} = \bm{f}_k\left(\bm{x}_k, \bm{u}_k^1, \dots, \bm{u}_k^N \right)\text{,}
\end{equation}
where $\bm{x}_k \in \mathcal{X} = \mathbb{R}^n$ and $\bm{u}_k^i \in \mathcal{U}^i = \mathbb{R}^{m}$. We consider $K$-stage games with an initial state $\bm{x}_0$, where the stage cost for player $i\in\mathcal{N}=\left\{1, 2, \dots, N\right\}$ depends on player $i$'s control inputs $\bm{u}_k^i \in \bm{u}^i=\left\{\bm{u}_0^i, \bm{u}_1^i, \dots, \bm{u}_{K-1}^i\right\}$ and the state $\bm{x}_k$. The sequence of states depends on $\bm{u}^i$ and the control inputs of all other players, which is often expressed with the index $-i$. Hence, the total cost of player $i$ depends on the initial state and all players' inputs:
\begin{align}
	\label{eq:game_nonlin_cost_input}
	J^i\left(\bm{x}_0, \bm{u}^i, \bm{u}^{-i}\right) = \sum_{k=0}^{K-1} g_k^i\left(\bm{x}_{k}, \bm{u}_k^i\right) +g_K^i\left(\bm{x}_{K}\right)\text{.}
\end{align}

A strategy $\gamma^i=\left\{\bm{\gamma}(\cdot)_0^i, \bm{\gamma}(\cdot)_1^i, \dots, \bm{\gamma}(\cdot)_{K-1}^i\right\}$ of the strategy space $\Gamma^i=\left\{\Gamma_0^i, \Gamma_1^i, \dots, \Gamma_{K-1}^i\right\}$ determines the control inputs for each stage $k$, depending on the available information to player $i$. The cost functional $\eqref{eq:game_nonlin_cost_input}$ expressed with strategies is:
\begin{equation}
	\label{eq:game_nonlin_cost_strategy}
	J^i\left(\bm{x}_0, \gamma^i,\gamma^{-i}\right) = \sum_{k=0}^{K-1} g_k^i\left(\bm{x}_{k}, \bm{\gamma}_k^i(\cdot)\right)+g_K^i\left(\bm{x}_{K}\right)\text{.}
\end{equation}
We omit the dependency on $\bm{x}_0$ for brevity in the following. An $N$-tuple of strategies $\left\{\gamma^{i*}\in\Gamma^i; i \in \mathcal{N}\right\}$ constitutes a Nash equilibrium if:

\begin{equation}
	\label{eq:nash_equilibrium}
	\forall i \in \mathcal{N}: J^i\left(\gamma^{i*}, \gamma^{-i*}\right) \le J^i\left(\gamma^{i}, \gamma^{-i*}\right)
\end{equation}
Loosely speaking, no player can improve its outcome at a Nash equilibrium by unilaterally altering its strategy.

The domain and codomain of the functions in the strategy space depend on the information structure of the game \cite{Basar1999}. The two information structures we consider lead to the following two  types of equilibria:
\subsubsection{Open-loop Nash equilibrium} 
In the open-loop case, all players observe the initial state $\bm{x}_0$ and generate a sequence of control inputs in a single act. This means, the strategy at stage $k$ in \eqref{eq:game_nonlin_cost_strategy} is a constant function with $\bm{\gamma}_k^i(\cdot) \in \Gamma_k^i=\mathcal{U}^i$. A Nash equilibrium $\left\{\gamma^{i*}\in\Gamma^i; i \in \mathcal{N}\right\}$ therefore directly translates to the players' input sequences $\left\{\bm{u}^{i*} = \gamma^{i*}; i \in \mathcal{N}\right\}$. A forward simulation of \eqref{eq:game_nonlin_dynamics} beginning with $\bm{x}_0^*=\bm{x}_0$ provides the corresponding open-loop state trajectory $\left\{\bm{x}_{k+1}^*; k\in\{0, 1, \dots, K-1\}\right\}$.

The open-loop problem in discrete time can be seen as a static infinite game, i.e., a game with infinite possible control input sequences of which one has to be chosen at the first and only stage $k=0$ \cite{Basar1999}.

\subsubsection{Feedback Nash equilibrium}
If the players know the current state $\bm{x}_k$, they can react to it and are not bound to an initially set sequence of control inputs. A feedback strategy $\gamma_k^i: \mathcal{X}\to\mathcal{U}^i$ maps the state to a control input $\bm{u}_k^i$ so that the control inputs at a stage $k$ corresponding to a Nash equilibrium are: $\left\{\bm{u}_k^{i*} = \gamma_k^{i*}(\bm{x}_k); i \in \mathcal{N}\right\}$. Such strategies can be calculated via dynamic programming, i.e., by working backward for $k$ from $K$ to $0$ and determining a Nash equilibrium for each static sub-game from stage $k$ to $k+1$.

In \acp{OCP}, which correspond to games with $N=1$ and only one cost-functional $J^1$, the trajectory obtained by simulating \eqref{eq:game_nonlin_dynamics} and applying the feedback solution in each stage coincides with the open-loop solution. This, however, does not apply to Nash equilibria of non-zero-sum games with $N>1$, even in the absence of disturbances or other unpredictable inputs. Starr and Ho \cite{Starr1969b} provide an illustrative example of this phenomenon and further examinations.

\section{Solving Discrete-Time Dynamic Games}
\label{sec:solving_game}
The \ac{iLQGame} approach in \cite{FridovichKeil2020} generates time-variant linear feedback laws for the players, yielding a feedback Nash equilibrium solution. However, \ac{iLQGame} can be adapted to generate an open-loop solution as done in the supplementary material of \cite{FridovichKeil2020}\footnote{\href{https://github.com/HJReachability/ilqgames}{https://github.com/HJReachability/ilqgames}}. In the following, we recapitulate the procedure which is summarized in Algorithm \ref{alg:iLQG}.
\begin{algorithm}[t]
	\label{alg:iLQG}
	\caption{\ac{iLQGame}}
	\begin{algorithmic}[1]
		\STATE \textbf{Input}: $\bm{x}_0$, initial trajectory $\hat{\bm{u}}^i$ and $\hat{\bm{x}}$ \label{alg:initial_guess}
		\STATE \textbf{Output}: Nash equilibrium trajectory $\bm{u}^{i*}$ and $\bm{x}^{*}$
		\WHILE{not converged}
		\FOR{$k\in\{0, 1, \dots, K\}$}
		\STATE $\bm{A}_k, \bm{B}_k^i \leftarrow$ \textsc{Linearize}$(\hat{\bm{x}}_k, \hat{\bm{u}}_k^1, \dots, \hat{\bm{u}}_k^N)$ \label{alg:linearize}
		\STATE $\bm{Q}_k^i, \bm{q}_k^i, \bm{R}_k^{ii}, \bm{r}_k^{ii} \leftarrow$ \textsc{Quadratize}$(\hat{\bm{x}}_k, \hat{\bm{u}}_k^1, \dots, \hat{\bm{u}}_k^N)$\label{alg:quadratize}	
		\ENDFOR
		\STATE $\bm{K}_k^i, \bm{k}_k^i \leftarrow$ \textsc{SolveLQGame}$(A_k, B_k^i, Q_k^i, q_k^i, R_k^{ii}, r_k^{ii})$\hspace{-1cm}\label{alg:solve_lq_game}
		\FOR{$k\in\{0, 1, \dots, K\}$}
		\STATE $\hat{\bm{u}}_k^{i,\mathrm{new}} \leftarrow$ \textsc{UpdateInput}$(\hat{\bm{x}}_k^\mathrm{new}, \bm{K}_k^i, \bm{k}_k^i)$ \label{alg:update_input}
		\STATE $\hat{\bm{x}}_{k+1}^{\mathrm{new}} = \bm{f}_k\left(\hat{\bm{x}}_k^{\mathrm{new}}, \hat{\bm{u}}_k^{1,\mathrm{new}}, \dots, \hat{\bm{u}}_k^{N,\mathrm{new}} \right)$ \label{alg:forward_pass}
		\ENDFOR
		\STATE $\hat{\bm{u}}^i \leftarrow \hat{\bm{u}}^{i, \mathrm{new}}$,  $\hat{\bm{x}} \leftarrow \hat{\bm{x}}^{\mathrm{new}}$
		\ENDWHILE
		\STATE$\bm{u}^{i*} \leftarrow \hat{\bm{u}}^i$,  $\bm{x}^{*}\leftarrow \hat{\bm{x}}$
	\end{algorithmic}
\end{algorithm}

Beginning with an initial state $\bm{x}_0$ and an initial guess for each player's control input sequence $\hat{\bm{u}}^i$, the initial nominal trajectory $\hat{\bm{x}}$ is obtained with \eqref{eq:game_nonlin_dynamics}.
\subsubsection{Linearization of the dynamics}\label{sec:linearization}
A linearization along the nominal trajectory provides the dynamic and input matrices for each time step and player. The resulting linear time-variant system is:
\begin{equation}
	\label{eq:game_linearized_dynamics}
	\begin{aligned}
		\Delta\bm{x}_{k+1} &= \bm{A}_k\Delta\bm{x}_{k} + \sum_{i=1}^{N}\bm{B}_k^i\Delta\bm{u}_{k}^i \text{ with }\\ \Delta\bm{x}_{k} &= \bm{x}_{k} - \hat{\bm{x}}_k \text{ and } \Delta\bm{u}_{k}^i = \bm{u}_{k} - \hat{\bm{u}}_k^i\text{.}
	\end{aligned}
\end{equation}
\subsubsection{Quadratization of the cost function}\label{sec:quadratization}
As in \ac{iLQR}, the stage cost is approximated by a second-order Taylor series:
\begin{equation}
	\label{eq:game_quadratized_cost}
	\begin{split}
		g_k^i&\left(\Delta\bm{x}_k, \Delta\bm{u}_k^1, \dots, \Delta\bm{u}_k^N\right)\thickapprox g_k^i\left(\hat{\bm{x}}_k, \hat{\bm{u}}_k^1, \dots, \hat{\bm{u}}_k^N\right) + \\&\qquad\underbrace{\left(\nabla_\mathbf{x}g_k^i|_{\hat{\bm{x}}_k,\hat{\bm{u}}_k^1,\dots,\hat{\bm{u}}_k^N}\right)^\top}_{\bm{q}_k^i} \Delta\bm{x}_{k} + 
		\\&\qquad\frac{1}{2}\Delta\bm{x}_{k}^\top\underbrace{\left(\nabla_\mathbf{x}^2 g_k^i|_{\hat{\bm{x}}_k,\hat{\bm{u}}_k^1,\dots,\hat{\bm{u}}_k^N}\right)}_{\bm{Q}_k^i}\Delta\bm{x}_{k} + 
		\\&\qquad\underbrace{\left(\nabla_{\mathbf{u}^i} g_k^i|_{\hat{\bm{x}}_k,\hat{\bm{u}}_k^1,\dots,\hat{\bm{u}}_k^N}\right)^\top}_{\bm{r}_k^{ii}} \Delta\bm{u}_{k}^i +
		\\&\qquad\frac{1}{2}\Delta\bm{u}_{k}^{i\top}\underbrace{\left(\nabla_{\mathbf{u}^i}^2 g_k^i|_{\hat{\bm{x}}_k,\hat{\bm{u}}_k^1,\dots,\hat{\bm{u}}_k^N}\right)}_{\bm{R}_k^{ii}}\Delta\bm{u}_{k}^i\text{.}
	\end{split}
\end{equation}
Here, we omit the mixed second-order terms since they do not appear in our cost function. Using the notation in \eqref{eq:game_quadratized_cost} and omitting the constant term, the total cost for player $i$ is approximated by:

\begin{equation}
	\label{eq:total_quad_cost}
	\begin{split}
		J^i \propto &\frac{1}{2}\sum_{k=0}^{K-1}\Biggl[\left(\Delta\bm{x}_k^\top \bm{Q}_k^i + 2\bm{q}_k^{i\top}\right)\Delta\bm{x}_k \Biggr.+\\ 
		&\qquad\qquad\Biggl.\sum_{j=1}^{N}\left[\left(\Delta\bm{u}_k^{j\top} \bm{R}_k^{ij} + 2\bm{r}^{ij\top}\right)\right]\Delta\bm{u}_k^j\Biggr] + \\
		& \frac{1}{2}\Delta\bm{x}_K^\top\bm{Q}_K^i\Delta\bm{x}_K + \bm{q}_K^{i\top}\Delta\bm{x}_K\text{.}
	\end{split}
\end{equation}
In our case, the mixed terms $\bm{R}_k^{ij}$ and $\bm{r}_k^{ij}$ with $i\neq j$ will be $0$. The cost structure \eqref{eq:total_quad_cost} and the linear dynamics \eqref{eq:game_linearized_dynamics} constitute a \ac{LQ} game that approximates the original game locally around the current nominal trajectory $\hat{\bm{x}}$.

\subsubsection{Solving the \ac{LQ} game}
For the \ac{LQ} game above, an analytical solution for the strategy of a feedback Nash equilibrium exists and has the linear affine form $\gamma_k^{i*}(\Delta\bm{x}_k) = -\bm{K}_k^i\Delta\bm{x}_k - \bm{k}_k^i$ \cite{Basar1999}. The elements in the matrices $\bm{K}_k^i$ and vectors $\bm{k}_k^i$ are obtained by solving the following systems of linear equations \cite{Basar1999, FridovichKeil2020}:
\begin{subequations}
	\label{eq:feedback_sol}
	\begin{align}
		&\begin{aligned}
			\left(\bm{R}_k^{ii} + \right. &\left.\bm{B}_k^{i\top}\bm{P}_{k+1}^i\bm{B}_k^i\right)\bm{K}_k^i + \bm{B}_k^{i\top}\bm{P}_{k+1}^i\cdot \\
			&\sum_{j=1, j\neq i}^{N}\left[\bm{B}_k^j\bm{K}_k^j\right] =\bm{B}_k^{i\top}\bm{P}_{k+1}^i\bm{A}_k
		\end{aligned}\label{eq:lin_sys_K}\\
		&\begin{aligned}\left(\bm{R}_k^{ii} + \right. &\left. \bm{B}_k^{i\top}\bm{P}_{k+1}^i\bm{B}_k^i\right)\bm{k}_k^i + \bm{B}_k^{i\top}\bm{P}_{k+1}^i\cdot \\
			&\sum_{j=1, j\neq i}^{N}\left[\bm{B}_k^j\bm{k}_k^j\right] =\bm{B}_k^{i\top}\bm{p}_{k+1}^i + \bm{r}_k^{ii}
		\end{aligned}
	\end{align}
\end{subequations}
As for linear-quadratic \acp{OCP} in discrete time, the matrices $\bm{P}_k^i$ and vectors $\bm{p}_k^i$ can be obtained by a recursion, which is given in Appendix \ref{sec:recursion_feedback}.

For the open-loop Nash equilibrium, the strategy $\gamma_k^{i*}(\cdot) = - \bm{k}_k^i$ does not depend on the current state and can be obtained with \cite{Basar1999, FridovichKeil2020}:
\begin{subequations}
	\begin{alignat}{2}
		&\bm{k}_k^{i} = -\bm{R}_k^{ii^{-1}}\left[\bm{B}_k^{i\top}\left(\bm{M}_{k+1}^i\Delta\bm{x}_{k+1} + \bm{m}_{k+1}^i\right) + \bm{r}_k^{ii}\right]\\
		&\begin{aligned}
			\Delta\bm{x}_{k+1} = \bm{\Lambda}_k^{-1}\Biggl[&\bm{A}_k\Delta\bm{x}_k - \Biggr. \\ &\Biggl.\sum_{j=1}^{N}\bm{B}_k^j\bm{R}_k^{jj^{-1}}\left(\bm{B}_k^{j\top}\bm{m}_{k+1}^j+\bm{r}_k^{jj}\right)\Biggr]\text{.}
		\end{aligned}
	\end{alignat}
\end{subequations}
Again, $\bm{M}_{k+1}^i$ and $\bm{m}_{k+1}^i$ are obtained by a recursion given in Appendix \ref{sec:recursion_open_loop}. Since both recursions proceed backward from $K$ to $0$, this step is often called the backward pass.

\subsubsection{Update Trajectory}
The forward pass updates the control inputs and nominal state trajectory according to the generated strategies. Due to the linearization of the dynamics and quadratization of the cost function along the trajectory, the obtained strategies are additive to the control inputs of the previous iteration. In the open-loop case, $\bm{K}_k^i$ is set to $\bm{0}$, and in the feedback case, it is applied on the difference from the previous iteration. Beginning with $\hat{\bm{x}}_0^{\mathrm{new}} = \hat{\bm{x}}_0 = \bm{x}_0$ the forward pass is:\\
For $k$ from $1$ to $K$:
\begin{subequations}
	\begin{alignat}{2}
		\hat{\bm{u}}_k^{i,\mathrm{new}} &= \hat{\bm{u}}_k^i - \bm{K}_k^i \left(\hat{\bm{x}}_k^{\mathrm{new}} - \hat{\bm{x}}_k\right) - \eta\bm{k}_k^i \\ \hat{\bm{x}}_{k+1}^{\mathrm{new}}&=\bm{f}_k\left(\hat{\bm{x}}_k^{\mathrm{new}}, \hat{\bm{u}}_k^{1, \mathrm{new}}, \dots, \hat{\bm{u}}_k^{N, \mathrm{new}}\right)\text{.}
	\end{alignat}
\end{subequations}
The scalar parameter $0<\eta\le 1$ can be interpreted as a step size and is usually chosen much smaller than $1$ to account for large deviations from the nominal trajectories where the approximations \eqref{eq:game_linearized_dynamics} and \eqref{eq:total_quad_cost} do not hold. With the new trajectory, the above sequence of linearization, quadratization, backward pass, and forward pass repeat until the algorithm converges.

Fridovich-Keil et al. \cite{FridovichKeil2020} point out that the resulting trajectory is not necessarily a Nash equilibrium of the original game. Instead, it represents a strategy that satisfies the conditions for a Nash equilibrium for a sequence of local approximations of the game.

\section{Racing Game}
\label{sec:dynamics_cost}
\subsection{Vehicle model and game dynamics}
Each player is modeled by a point mass following \cite{Rowold.2023} where the state includes the progress $s$, velocity $V$, lateral displacement $n$, relative orientation $\chi$ towards the track's reference line with the curvature $\kappa(s)$, and the longitudinal and lateral accelerations $a_\mathrm{x}$ and $a_\mathrm{y}$. The control input vector includes the jerks in longitudinal and lateral directions: $\bm{u}^\top = \begin{bmatrix}
	j_\mathrm{x} & j_\mathrm{y}
\end{bmatrix}$. The time-continuous nonlinear dynamics of player $i$ are given by:
\begin{equation}
	\label{eq:dynamics}
	\dot{\mathbf{x}}^i = \begin{bmatrix}
		\dot{s}^i\\
		\dot{V}^i\\
		\dot{n}^i\\
		\dot{\chi}^i\\
		\dot{a}_\mathrm{x}^i\\
		\dot{a}_\mathrm{y}^i
	\end{bmatrix}=\tilde{\mathbf{f}}^i(\mathbf{x}^i, \mathbf{u}^i) = \begin{bmatrix}
		\frac{V^i\cos(\chi^i)}{1-n^i\kappa(s^i)}\\
		a_\mathrm{x}^i\\
		V^i\sin(\chi^i) \\
		\frac{a_\mathrm{y}^i}{V^i}-\kappa(s^i)\frac{V^i\cos(\chi^i)}{1-n^i\kappa(s^i)}\\
		j_\mathrm{x}^i\\
		j_\mathrm{y}^i
	\end{bmatrix}\text{.}
\end{equation}
Since racing cars often operate at the handling limits, it is important to constrain the accelerations to obtain feasible trajectories. Similar to \cite{Rowold.2023}, we approximate the velocity-dependent gg-diagrams by diamonds with a maximum positive acceleration $a_\mathrm{x}\le a_\mathrm{x, max}(V)$ and a maximum combined radius $\rho(V)$:
\begin{equation}
	\sqrt{a_\mathrm{x}^2 + a_\mathrm{y}^2} \le \rho(V)\text{.}
\end{equation}

The joint state vector of the game is a concatenation of $N$ player state vectors:
\begin{equation}
	\dot{\bm{x}} = \begin{bmatrix}
		\dot{\bm{x}}^1\\
		\vdots\\
		\dot{\bm{x}}^N
	\end{bmatrix} = \begin{bmatrix}
		\tilde{\mathbf{f}}^1(\mathbf{x}, \mathbf{u}^i)\\
		\vdots\\
		\tilde{\mathbf{f}}^N(\mathbf{x}, \mathbf{u}^N)
	\end{bmatrix} = \tilde{\bm{f}}(\bm{x}, \bm{u}^1, \dots, \bm{u}^N)
\end{equation}

\subsection{Cost function}
\ac{LQR} approaches naturally do not consider state and input constraints. Chen et al. \cite{Chen2017} realize constraints in \ac{iLQR} through the cost function and introduce barrier functions. Quadratic cost terms for constraint violations in \cite{FridovichKeil2020} show good results regarding convergence and robustness of the \ac{iLQGame} algorithm. Our stage costs, including the constraints, are:
\begin{subequations}
	\begin{alignat}{2}
		g&_k^i =\bm{u}_k^{i\top}\bm{R}^i\bm{u}_k^i+\label{eq:regularization}\\
		&\sum_{j=1, j\neq i}^{N}c_\mathrm{c}^i\left(e^{1-\left(\frac{s_k^i -s_k^j}{l_\mathrm{veh}}\right)^2 - \left(\frac{n_k^i -n_k^j}{w_\mathrm{veh}}\right)^2}\right)^2\label{eq:collision_cost} +\\
		&\bm{1}\left\{n_k^i\ge w_\mathrm{tr, l/r}(s_k^i) \right\}c_\mathrm{w}^i\left(n_k^i-w_\mathrm{tr, l/r}(s_k^i)\right)^2 + \label{eq:track_cost}\\
		&\bm{1}\left\{a_{\mathrm{x},k}^i\ge a_\mathrm{x, max}(V_k^i)\right\}c_{a_\mathrm{x}}^i\left(a_{\mathrm{x}, k}^i- a_\mathrm{x, max}(V_k^i)\right)^2 + \\
		&\bm{1}\left\{\sqrt{a_\mathrm{x}^2 + a_\mathrm{y}^2} \ge \rho(V)\right\}c_a^i\left(\sqrt{a_\mathrm{x}^2 + a_\mathrm{y}^2} - \rho(V)\right)^2\text{.}
	\end{alignat}
\end{subequations}
As in \cite{FridovichKeil2020}, we use the operator $\bm{1}\{\cdot\}$ which becomes $1$ if the condition holds, and $0$ otherwise. The term \eqref{eq:regularization} regularizes the jerk as in \cite{Rowold.2023} and \eqref{eq:collision_cost} introduces a coupling between players by penalizing collisions. As player $i$ and $j$ come closer, the term increases with longitudinal and lateral distances weighted differently.

All other stage cost terms implement soft constraints with the weights $c_\mathrm{w}$, $c_{a_\mathrm{x}}$, $c_a$, $c_\mathrm{c}$. \eqref{eq:track_cost} enforces the track boundaries with the track widths to the left and right $w_\mathrm{tr, l/r}$. Note that none of the state constraints depend on the inputs in $\bm{u}_k^i$, as this would result in mixed second-order terms in \eqref{eq:game_quadratized_cost}. 

The terminal costs introduce another coupling and should provide the incentive to drive fast and to be ahead at the end of the planning horizon:
\begin{equation}
	g_K^i = -s_K^i + c_\mathrm{g}^i \sum_{j=1, j\neq i}^{N}s_K^j
\end{equation}
The first term penalizes little progress and the second term with the weight $c_\mathrm{g}$ should incentivize defending or blocking maneuvers. A similar terminal cost for racing is used in \cite{Wang2021}.

\section{Exemplary Results}
\label{sec:results}
\begin{figure}
	\centering
	\def\axiswidth{8.5cm}
	\def\vertdis{0.5cm}
	\input{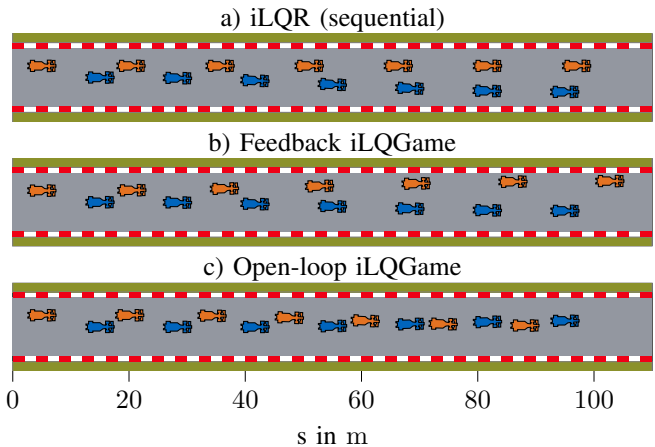}
	\vspace{-0.75cm}
	\caption{Results for \ac{iLQR}, open-loop \ac{iLQGame}, and feedback \ac{iLQGame}: All three methods use the same cost parameters and are initialized with $\hat{\bm{u}}^i=\bm{0}$.}
	\label{fig:1}
\end{figure}

This section provides examples to demonstrate the capability of \ac{iLQGame} to consider interactions in racing scenarios and to motivate comparing the two types of Nash equilibria. The considered scenario in Figures \ref{fig:1} and \ref{fig:2} includes the ego vehicle ($i=1$, blue) with a maximum velocity of \SI{20}{\meter\per\second} and the opponent ($i=2$, orange) with \SI{25}{\meter\per\second}. The opponent approaches the ego vehicle with \SI{23}{\meter\per\second} and a lateral displacement of \SI{2}{\meter}. For the following results, we initialize the control input sequences with $\hat{\bm{u}}^i=\bm{0}$.

Figure \ref{fig:1} a) shows the trajectories obtained with a sequential approach. The opponent vehicle is predicted assuming a constant velocity and lateral displacement. With the fixed prediction, the \ac{iLQGame} algorithm reduces to \ac{iLQR}, and the resulting trajectory swerves to the right to avoid collisions. This scenario highlights the importance of interaction-aware planning since the observed yielding behavior is not desirable in competitive racing.

The feedback solution is shown in Figure \ref{fig:1} b). The right swerving maneuver of the ego vehicle occurs to a lesser degree due to its awareness that the opponent is also trying to avoid collisions. Increasing the collision cost weight of the opponent $c_\mathrm{c}^{i=2}$ as shown in Figure \ref{fig:2} results in a greater leveraging of the opponent's reaction so that the ego vehicle can maintain its course. The choice of $c_\mathrm{c}^{i=2}>c_\mathrm{c}^{i=1}$ can be justified assuming that the trailing vehicle bears a greater responsibility to avoid collisions.

Figure \ref{fig:1} c) shows the trajectories of the open-loop solution. The players' behaviors significantly differ from the feedback solution, and the ego vehicle performs a blocking maneuver. However, we want to point out that when the open-loop \ac{iLQGame} algorithm is initialized with the feedback solution, it terminates after the first iteration, yielding the same outcome.

\section{Outlook}
\begin{figure}
	\centering
	\def\axiswidth{8.5cm}
	\def\vertdis{0.5cm}
	\input{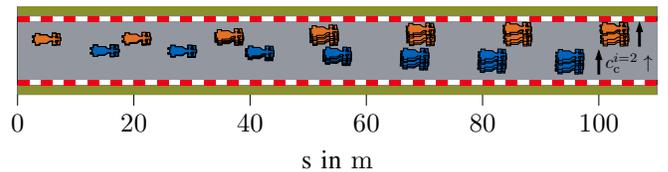}
	\vspace{-0.75cm}
	\caption{Variation of $c_\mathrm{c}^{i=2}$ for feedback \ac{iLQGame}: A greater ratio of $\nicefrac{c_\mathrm{c}^{i=2}}{c_\mathrm{c}^{i=1}}$ causes the ego vehicle to be more aggressive, forcing the opponent to swerve further to the left.}
	\label{fig:2}
\end{figure}
The examples in Section \ref{sec:results} illustrate \ac{iLQGame}'s capability to consider interactions in racing scenarios. The algorithm converges to different solutions in the open-loop and feedback cases when initialized with identical input sequences. However, when initialized differently, both concepts can yield the same solution. This phenomenon is consistent with the non-uniqueness of Nash equilibria in dynamic nonzero-sum games. Therefore, the convergence property of \ac{iLQGame} should be further examined in future work. The investigation should include the influence of the initialization and of the step size $\eta$.

Our current analysis is limited to one planning step, whereas planning algorithms are usually applied with a moving horizon. The algorithm's initialization is then based on the solution from the previous planning step. Future work should assess the outcomes regarding performance and safety when the two types of equilibria are used with a moving horizon. The analyses should also consider more complex race tracks and the case in which the opponent employs a sequential approach to evaluate the robustness when exposed to a non-interaction-aware player. Ultimately, the analyses should conclude whether \ac{iLQGame} is suited for racing scenarios and whether an open-loop or a feedback solution should be preferred.

As indicated in Figure \ref{fig:2}, the cost parameterization influences the ego vehicle's aggressiveness. Further experiments should determine reasonable racing parameterizations and identify possibly online adjustable parameters to gain an advantage during a race while maintaining safe behaviors. These parameters may depend, e.g., on the current position relative to the opponent, i.e., whether the ego vehicle is leading or trailing.

\bibliographystyle{IEEEtran}
\bibliography{references}

\section*{APPENDIX}
\label{sec:appendix}
The derivations of the following recursions \eqref{eq:feedback_recursion} and \eqref{eq:open_loop_recursion} without linear cost terms are given in \cite{Basar1999}. The supplementary material to \cite{FridovichKeil2020} (\href{https://github.com/HJReachability/ilqgames/tree/master/derivations}{https://github.com/HJReachability/ilqgames/tree/master/\ derivations}) provides the extensions with linear cost terms.

\subsection{Recursion for the feedback equilibrium}
\label{sec:recursion_feedback}
Beginning with $\bm{P}_K^i = \bm{Q}_K^i$ and $\bm{p}_K^i = \bm{q}_K^i$:\\
For $k$ from $K-1$ to $0$:
\begin{subequations}
	\label{eq:feedback_recursion}\\
	\begin{alignat}{2}
		& \bm{F}_k = \bm{A}_k - \sum_{j=1}^{N}\bm{B}_k^j\bm{K}_k^j \text{, \quad} \bm{\beta}_k = - \sum_{j=1}^{N}\bm{B}_k^j\bm{k}_k^j\\
		& \bm{P}_k^i = \bm{Q}_k^i + \bm{F}_k^\top\bm{P}_{k+1}^i\bm{F}_k + \sum_{j=1}^{N}\bm{K}_k^{j\top}\bm{R}_k^{ij}\bm{K}_k^j\label{eq:coupled_riccati}\\
		&\begin{aligned}
			\bm{p}_k^i = \bm{q}_k^i + &\bm{F}_k^\top\left(\bm{p_{k+1}^i} + \bm{P}_{k+1}^i\bm{\beta}_k\right) + \\ &\sum_{j=1}^{N}\left[\bm{K}_k^{j\top}\bm{R}_k^{ij}\bm{k}_k^j - \bm{K}_k^{j\top}\bm{r}_k^{ij}\right]\text{.}
		\end{aligned}
	\end{alignat}
\end{subequations}
It should be noted, that for $N=1$ and $\bm{q}_k = \bm{r}_k = 0$, \eqref{eq:coupled_riccati} together with \eqref{eq:lin_sys_K} simplify to the \emph{difference Riccati equation} (dropping index $i$):
\begin{equation}
	\begin{aligned}
		\bm{P}_k = \bm{Q}_k + &\bm{A}_k^\top\bm{P}_{k+1}\bm{A}_k - \left(\bm{A}_k^\top\bm{P}_{k+1}\bm{B}_k\right)\cdot\\&\left(\bm{R}_k+\bm{B}_k^\top\bm{P}_{k+1}\bm{B}_k\right)^{-1}\left(\bm{B}_k^\top\bm{P}_{k+1}\bm{A}_k\right)\text{,}
	\end{aligned}
\end{equation}
which is well known from \acp{LQR} in discrete time.

\subsection{Recursion for the open-loop equilibrium}
\label{sec:recursion_open_loop}
Beginning with $\bm{M}_K^i = \bm{Q}_K^i$ and $\bm{m}_K^i = \bm{q}_K^i$:\\
For $k$ from $K-1$ to $0$:
\begin{subequations}
	\label{eq:open_loop_recursion}\\
	\begin{alignat}{2}
		&\bm{\Lambda}_k = \mathbb{I} + \sum_{j=1}^{N}\bm{B}_k^j\bm{R}_k^{jj^{-1}}\bm{B}_k^{j\top}\bm{M}_{k+1}^j\\
		&\begin{aligned}
			\bm{m}_k^i = \bm{A}_k^\top\Biggl[&\bm{m}_{k+1}^i-\bm{M}_{k+1}^i\bm{\Lambda}_k^{-1}\cdot\Biggr.\\
			\Biggl.&\sum_{j=1}^{N}\bm{B}_k^j\bm{R}_k^{jj^{-1}}\left(\bm{B}_k^{j\top}\bm{m}_{k+1}^j+\bm{r}_k^{jj}\right)\Biggr]+\bm{q}_k^i
		\end{aligned}\\
		&\bm{M}_k^i = \bm{Q}_k^i+\bm{A}_k^\top\bm{M}_{k+1}^i\bm{\Lambda}_k^{-1}\bm{A}_k\text{.}
	\end{alignat}
\end{subequations}

\end{document}

%% file: custom.tex
\usepackage{hyperref}
\usepackage{acro}
\acsetup{
	single-style = long,
	single = true
}
\usepackage{algorithm}
\usepackage{xcolor}
\usepackage{siunitx}
\usepackage{nicefrac}
\usepackage{tikz}
\usepackage[utf8]{inputenc}
\usepackage{pgfplots}
\DeclareUnicodeCharacter{2212}{−}
\usepgfplotslibrary{groupplots,dateplot}
\usetikzlibrary{patterns,shapes.arrows}
\pgfplotsset{compat=newest}
\usepackage{cite}
\usepackage{amsmath,amssymb,amsfonts}
\usepackage[noend]{algorithmic}
\usepackage{graphicx}
\usepackage{textcomp}
\usepackage{xcolor}
\usepackage{bm}

\DeclareAcronym{LQR}{
	short=LQR,
	long=linear-quadratic regulator
}

\DeclareAcronym{iLQR}{
	short=iLQR,
	long=iterative linear-quadratic regulator
}

\DeclareAcronym{iLQG}{
	short=iLQG,
	long=iterative linear-quadratic Gaussian
}

\DeclareAcronym{LQ}{
	short=LQ,
	long=linear-quadratic
}

\DeclareAcronym{CFR}{
	short=CFR,
	long=counterfactual regret
}

\DeclareAcronym{iLQGame}{
	short=iLQGame,
	long=iterative linear-quadratic game
}

\DeclareAcronym{OCP}{
	short=OCP,
	long=optimal control problem
}

\DeclareAcronym{POMDP}{
	short=POMDP,
	long=partially observable Markov desicion processes
}

\DeclareAcronym{IBR}{
	short=IBR,
	long=iterative best response
}

\DeclareAcronym{DDP}{
	short=DDP,
	long=differential dynamic programming
}

\DeclareAcronym{SQP}{
	short=SQP,
	long=sequential quadratic programming
}

\DeclareAcronym{KKT}{
	short=KKT,
	long=Karush–Kuhn–Tucker,
}